\documentclass[a4paper,fleqn, review]{cas-dc}
\usepackage{amsmath}
\usepackage[switch]{lineno}
\usepackage{graphicx}
\usepackage{subfig} 
\usepackage{tabu}  
\usepackage{hyperref}
\usepackage{multirow}
\usepackage{dsfont}
\usepackage{stfloats}
\usepackage{cite}
\usepackage{booktabs}
\usepackage{array} 
\usepackage{algorithm}
\usepackage{amssymb}
\usepackage{tikz}
\usepackage{makecell}
\usepackage{caption}
\usepackage{natbib}
\usepackage{siunitx}
\usepackage{pifont}
\usepackage{booktabs}
\usepackage{tabularx}
\usepackage{array}
\usepackage{graphics}
\usepackage[export]{adjustbox}
\usepackage{algorithm}
\usepackage{algpseudocode}
\usepackage{xcolor}
\def\tsc#1{\csdef{#1}{\textsc{\lowercase{#1}}\xspace}}
\tsc{WGM}
\tsc{QE}
\tsc{EP}
\tsc{PMS}
\tsc{BEC}
\tsc{DE}

\definecolor{bareland_color}{rgb}{0.502,0.0,0.0}
\definecolor{cropland_color}{rgb}{0.655, 0.733,0.106}
\definecolor{vegetation_color}{rgb}{0.274,0.710,0.474}
\definecolor{water_color}{rgb}{0.110,0.549,0.741}
\definecolor{building_color}{rgb}{0.710, 0.274, 0.274}
\definecolor{road_color}{rgb}{0.870,0.721,0.274}
\definecolor{developed_space_color}{rgb}{0.580,0.580,0.580}
\definecolor{background_color}{rgb}{0.949,0.937,0.914}

\begin{document}
\let\WriteBookmarks\relax
\def\floatpagepagefraction{1}
\def\textpagefraction{.001}
\shorttitle{}

\shortauthors{J, Song et~al.}

\title [mode = title]{RS-MTDF: Multi-Teacher Distillation and Fusion for Remote Sensing Semi-Supervised Semantic Segmentation}



%

\author[1]{Jiayi Song$^{*}$}[orcid=0009-0008-4585-4735]

\author[2]{Kaiyu Li$^{*}$}[orcid=0000-0002-8015-6245]

\author[1,3]{Xiangyong Cao$^{\dag}$}

\author[4, 5]{Deyu Meng}





\affiliation[1]{organization={School of Computer Science and Technology, Xi’an Jiaotong University},
    city={Xi’an},
    postcode={710049}, 
    country={China}}

\affiliation[2]{organization={School of Software Engineering, Xi’an Jiaotong University},
    city={Xi’an},
    postcode={710049}, 
    country={China}}


\affiliation[3]{organization={Ministry of Education Key Lab of Intelligent Network Security, Xi’an Jiaotong University},
    city={Xi’an},
    postcode={710049}, 
    country={China}}
    
\affiliation[4]{organization={School of Mathematics and Statistics, Xi’an Jiaotong University},
    city={Xi’an},
    postcode={710049}, 
    country={China}}

\affiliation[5]{organization={Pazhou Laboratory (Huangpu)},
    addressline={Radarweg 29}, 
   city={Guangzhou},
    citysep={}, 
   postcode={510555}, 
    state={},
   country={China}}




\cortext[cor1]{Equal contribution. \dag Corresponding author.}


\nonumnote{E-mail addresses: mailto:songyangyifei@gmail.com (J, Song), likyoo.ai@gmail.com (K. Li), caoxiangyong@mail.xjtu.edu.cn (X. Cao), dymeng@mail.xjtu.edu.cn (D. Meng)}

\begin{abstract}
Semantic segmentation in remote sensing images is crucial for various applications, yet its performance is heavily reliant on large-scale, high-quality pixel-wise annotations, which are notoriously expensive and time-consuming to acquire. Semi-supervised semantic segmentation (SSS) offers a promising alternative to mitigate this data dependency. However, existing SSS methods often struggle with the inherent distribution mismatch between limited labeled data and abundant unlabeled data, leading to suboptimal generalization. To alleviate this issue, we attempt to introduce the Vision Foundation Models (VFMs) pre-trained on vast and diverse datasets into the SSS task since VFMs possess robust generalization capabilities that can effectively bridge this distribution gap and provide strong semantic priors for SSS. Inspired by this, we introduce RS-MTDF (Multi-Teacher Distillation and Fusion), a novel framework that leverages the powerful semantic knowledge embedded in VFMs to guide semi-supervised learning in remote sensing. Specifically, RS-MTDF employs multiple frozen VFMs (\textit{e.g.}, DINOv2 and CLIP) as expert teachers, utilizing feature-level distillation to align student features with their robust representations. To further enhance discriminative power, the distilled knowledge is seamlessly fused into the student decoder. Extensive experiments on three challenging remote sensing datasets (ISPRS Potsdam, LoveDA, and DeepGlobe) demonstrate that RS-MTDF consistently achieves state-of-the-art performance. Notably, our method outperforms existing approaches across various label ratios on LoveDA and secures the highest IoU in the majority of semantic categories. These results underscore the efficacy of multi-teacher VFM guidance in significantly enhancing both generalization and semantic understanding for remote sensing segmentation. Ablation studies further validate the contribution of each proposed module. Code is available at \url{https://github.com/earth-insights/RS-MTDF}.
\end{abstract}


\maketitle

\section{Introduction}\label{sec:1}
Leveraging deep learning techniques, remote sensing image semantic segmentation has emerged as a critical solution for various applications, such as urban planning, land use monitoring, disaster management, and environmental protection \citep{swin,li2024new, li2024open}. With significant advancements in remote sensing and earth observation technologies, the availability of high-resolution remote sensing images is rapidly increasing, offering unprecedented opportunities for more accurate and automated analysis. However, traditional supervised semantic segmentation \citep{guo2018review} heavily relies on large-scale, high-quality pixel-wise annotations \citep{li2025segearth}. Compared to natural images, remote sensing images often contain multi-scale objects, complex boundaries, and diverse geographical features, making high-quality manual annotation notoriously expensive, laborious, and time-consuming. To address this critical limitation, semi-supervised semantic segmentation (SSS) \citep{surveysemisupervisedsemanticsegmentation} emerges as a viable paradigm, enabling models to be trained effectively with a small amount of labeled data complemented by a large volume of unlabeled data.

\begin{figure}[t]
  \centering
   \includegraphics[width=1.0\linewidth]{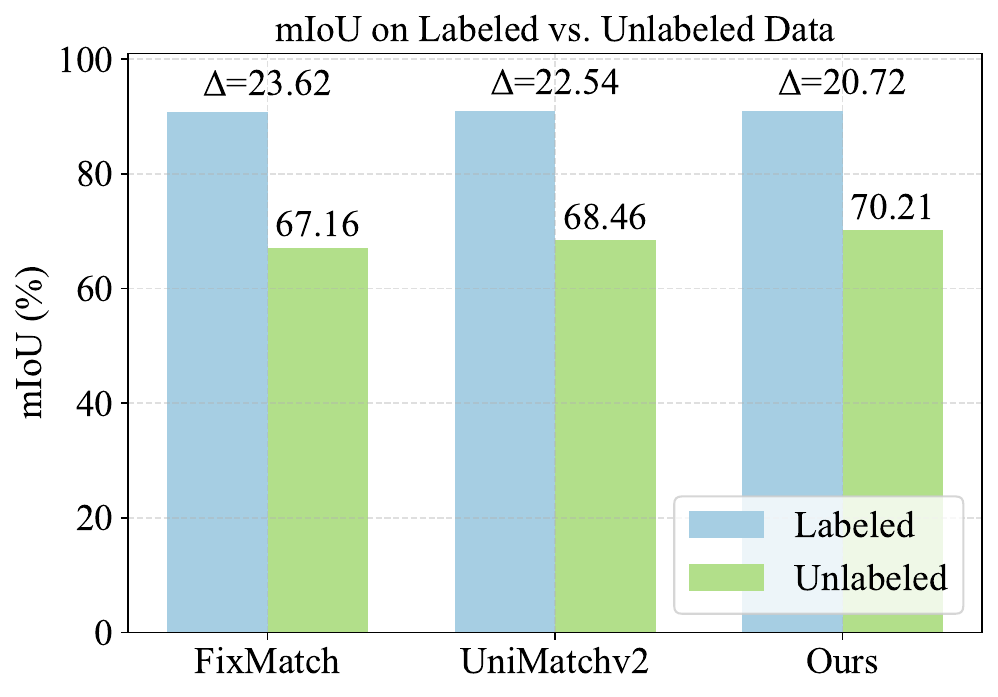}
   \caption{This figure illustrates the performance gap in mIoU between labeled and unlabeled data across FixMatch~\citep{fixmatch}, UniMatchv2~\citep{unimatchv2}, and our proposed method. Both FixMatch and UniMatchv2 exhibit a large discrepancy between the two data types, indicating limited generalization to unlabeled samples. In contrast, our method effectively reduces this gap while also achieving a notable improvement in mIoU on the unlabeled set, demonstrating its overall effectiveness.}
   \label{fig:fig1}
\end{figure}


Despite promising advances in SSS for remote sensing \citep{dwl,semisupervisedsemanticsegmentationremote}, a critical, often underestimated, challenge persists: the inherent distribution mismatch between limited labeled data and abundant unlabeled data. While most existing SSS methods implicitly assume that labeled and unlabeled data originate from the same underlying distribution, this assumption rarely holds true in real-world remote sensing scenarios. In practice, labeled and unlabeled data often exhibits significant domain shifts due to varying sensor types, acquisition times, geographical regions, or atmospheric conditions. Under current semi-supervised frameworks, the learning processes for labeled and unlabeled data are tightly coupled. This tight coupling makes it difficult for the model to balance attention between the two, especially when their distributions differ. As shown in Figure~\ref{fig:fig1}, the mIoU performance on labeled and unlabeled samples diverges significantly, indicating that the model tends to overfit to the labeled data and struggles to generalize effectively to the unlabeled set.

Benefiting from large-scale pretraining on vast and diverse visual data, Vision Foundation Models (VFMs) possess robust generalization capabilities and rich semantic priors \citep{clip, dino}, making them ideal for bridging this distribution gap. This motivates us to leverage VFMs as powerful guidance sources to address the aforementioned challenges in semi-supervised remote sensing segmentation. We propose RS-MTDF (Multi-Teacher Distillation and Fusion), a novel framework that synergistically leverages multiple frozen VFMs as expert teachers to guide the student model in learning more generalizable representations. Specifically, we select DINOv2 \citep{dino} for its robust local feature extraction and spatial sensitivity \citep{wang2022self}, and CLIP \citep{clip} for its global contextual understanding and broad semantic priors derived from vision-language alignment \citep{wysoczanska2024clip}. By performing feature-level distillation \citep{heo2019comprehensive} at the encoder stage, we transfer the rich knowledge embedded in these pretrained models to the student encoder, thereby imparting richer, more robust representations. This knowledge transfer enables the student model to better capture complex spatial and semantic patterns in remote sensing image, significantly enhancing its generalization ability and robustness under low-label conditions.

To further exploit the deep knowledge embedded in teacher encoder representations, we incorporate the translated student features into the decoding stage. Specifically, we fuse these projected features with the student's high-level representation before passing them into the segmentation head. This integrated fusion mechanism ensures that the distilled knowledge not only influences early feature learning but also refines fine-grained segmentation details at later stages, leading to more accurate and discriminative predictions.

Our main contributions are summarized as follows:
\begin{enumerate}
    \item To address the critical and persistent challenge of distribution mismatch between labeled and unlabeled data in semi-supervised remote sensing segmentation, we propose RS-MTDF, a novel framework that explicitly and effectively alleviates this problem.
    
    \item We introduce a multi-teacher distillation strategy, where VFMs guide the student encoder to learn more generalizable representations. To further exploit the transferred knowledge, we fuse the translated student features with the original student representations before feeding them into the decoder.
    
    \item We conduct extensive experiments on three challenging remote sensing datasets (ISPRS Potsdam, LoveDA, and DeepGlobe), demonstrating that our approach consistently achieves state-of-the-art performance, particularly in low-label conditions. Ablation studies further validate the critical contribution of each proposed module.
\end{enumerate}

The remainder of this paper is organized as follows. Section~\ref{sec:rw} provides a brief overview of SSS, including its application in remote sensing and the associated challenges. Section~\ref{sec:method} describes the proposed semi-supervised framework in detail. Extensive experimental results and in-depth discussions are presented in Section~\ref{sec:exp}. Section~\ref{sec:conclusion} concludes the paper. Finally, the limitations of the proposed method and directions for future work are discussed in Section~\ref{sec:limitation}.


\section{Related Works}
\label{sec:rw}
\subsection{Semi-supervised Semantic Segmentation}

The central challenge in SSS lies in effectively leveraging the vast amount of unlabeled data to augment performance achieved with limited labeled examples. Conventionally, two principal paradigms have dominated SSS research: pseudo-labeling and consistency regularization.

Pseudo-labeling \citep{Pseudo-labelin}, a cornerstone of self-training, operates by iteratively assigning ``pseudo labels'' to unlabeled samples based on the model's predictions from previous training iterations. These high-confidence pseudo labels then serve as additional supervision signals, expanding the training set. Consistency regularization \citep{consistencyregularization}, conversely, is premised on the ``smoothness assumption'' \citep{surveysemisupervisedsemanticsegmentation}: predictions for the same unlabeled input should remain consistent under various perturbations. This is typically enforced by minimizing a consistency loss between predictions from different augmented views or perturbed models of the same unlabeled data.

In recent years, many state-of-the-art SSS methods have successfully integrated both pseudo-labeling and consistency regularization \citep{unimatch,surveysemisupervisedsemanticsegmentation}. FixMatch \citep{fixmatch} has emerged as a de-facto baseline, built upon the mean-teacher framework \citep{meanteachersbetterrole}. In FixMatch, a teacher model generates high-confidence pseudo-labels for weakly augmented unlabeled images. These pseudo-labels then supervise the student model's predictions on strongly augmented versions of the same images, enforcing consistency and leveraging implicit data augmentation.

Building upon FixMatch, subsequent research has explored various avenues to enhance SSS. Some works focus on expanding the perturbation space for consistency regularization. For example, UniMatch \citep{unimatch} extends beyond image-level perturbations to include feature-level perturbations, aiming for more robust learning. AugSeg \citep{augmentationmatterssimpleyeteffectiveapproach} emphasizes diverse data augmentation strategies, generating multiple distinct views of the same image to maximize consistency benefits. Other studies concentrate on improving the quality and reliability of pseudo labels. \cite{sun2023daw} argued that solely relying on high-confidence predictions might neglect valid but low-confidence labels, proposing Daw to address this. CorrMatch \citep{corrmatchlabelpropagationcorrelation} further refines pseudo-label generation by incorporating spatial correlation maps, enhancing the reliability of generated labels by considering contextual information.

\subsection{Semi-Supervised Semantic Segmentation in Remote Sensing Images}


Semi-supervised semantic segmentation has gained increasing traction in remote sensing images, driven by the intrinsic characteristics of remote sensing data, such as its high resolution, multi-scale objects, complex boundaries, and frequent inter-class similarities. These properties significantly exacerbate the annotation burden, making SSS a particularly appealing solution.

Early efforts adapted general SSS techniques to remote sensing. \cite{he2022classhyper} proposed a hybrid consistency regularization framework that combined data-level and model-level perturbations, specifically incorporating semantic boundary information during data mixing to enhance spatial precision, a crucial aspect for remote sensing data. \cite{icnetrealtimesemanticsegmentation} introduced ICNet, which leveraged a teacher-student framework to improve pseudo-label quality and employed an iterative training strategy to increase model diversity. More recently, \cite{dwl} proposed a decoupled weighted learning (DWL) framework, introducing a decoupling module to separate the training processes of labeled and unlabeled data. This aims to mitigate the adverse impact of noisy pseudo-labels, which are more prevalent in challenging remote sensing environments. Addressing the complexities of rich multi-scale features and high inter-class similarity, MUCA \citep{semisupervisedsemanticsegmentationremote} was proposed by Wang \textit{et al.}, enforcing consistency across feature maps from different network layers via multi-scale uncertainty-aware consistency regularization.

\subsection{Challenges in Remote Sensing SSS and Some Potential Solutions}

Despite the aforementioned advances, a significant challenge remains: the inherent distributional discrepancies between limited labeled data and abundant unlabeled data. As highlighted in Section~\ref{sec:1}, FixMatch-based methods and their derivatives often struggle to bridge this gap, leading to models that overfit labeled data and exhibit suboptimal generalization to the unlabeled set. Existing remote sensing SSS methods typically assume that labeled and unlabeled data come from the same distribution, an assumption rarely met in diverse real-world remote sensing scenarios (\textit{e.g.}, due to different sensors, acquisition times, or geographic regions). To address this, some studies have explored leveraging global context and more robust feature learning. AllSpark \citep{allsparkrebornlabeledfeatures}, for instance, utilizes the global feature extraction capability of Transformers to learn more label-relevant representations from unlabeled data. BCP \citep{BCP} introduced a bi-directional copy-pasting mechanism between labeled and unlabeled images to promote stronger semantic fusion during training, aiming to bridge data disparities.

In parallel, the emergence of VFMs, pre-trained on vast and diverse visual datasets, has demonstrated unparalleled generalization capabilities beyond their training domains. This motivates their application as powerful sources of rich semantic and spatial priors for SSS in remote sensing. Instead of traditional CNN-based backbones \citep{deeplabsemanticimagesegmentation} like ResNet \citep{deepresiduallearningimage}, VFMs offer a new paradigm. For example, SemiVL \citep{semivl} integrates rich priors from CLIP \citep{clip} pretraining to enhance semantic boundary learning, crucial for precise segmentation. However, SemiVL primarily leverages CLIP for direct pseudo-label generation, which can be sensitive to the quality of CLIP's raw predictions and may lack fine-grained pixel-level understanding necessary for precise segmentation in remote sensing images~\citep{semicdvlvisuallanguagemodelguidance}. Similarly, UniMatchv2 \citep{unimatchv2} employs DINO \citep{dino} as its backbone, capitalizing on DINO's strong spatial awareness acquired through self-supervised learning on large datasets. While beneficial as a strong initialization, merely utilizing DINO's weights as a pre-trained backbone risks diluting its inherent strong generalization capabilities over the course of training, as the model adapts to the specific labeled data and may fail to fully leverage the broad semantic priors from the VFM. These works undeniably highlight the potential of integrating large-scale pre-trained models to enhance SSS performance, particularly in scenarios where data distribution shifts are prominent. Nevertheless, their reliance on a single VFM and specific integration strategies might restrict the full exploitation of VFM knowledge and their generalization capabilities across various RS image characteristics. 

This indicates a critical need for a more robust and adaptive mechanism to leverage VFM knowledge effectively. Our work addresses these limitations by proposing RS-MTDF, a novel framework that goes beyond single VFM integration. We employ multiple powerful VFMs (DINOv2 and CLIP) as expert teachers, not merely for initialization or pseudo-label generation, but through a dedicated feature-level distillation process. This multi-teacher approach allows us to robustly transfer and fuse complementary semantic and spatial priors, ensuring the student model retains and fully exploits the VFMs' generalization power, thereby alleviating the distribution mismatch between labeled and unlabeled data in remote sensing segmentation.



\section{Method}
\label{sec:method}

In this section, we introduce RS-MTDF, a novel semi-supervised semantic segmentation framework specifically designed for remote sensing images. Our method builds upon the well-established FixMatch framework due to its robust pseudo-labeling and consistency regularization mechanisms. Therefore, this section first provides a brief re-introduction to the FixMatch framework in Section~\ref{sec:Preliminaries}, outlining its core components and loss functions. Subsequently, Section~\ref{sec:3.2} details our proposed RS-MTDF, describing its architectural innovations, multi-teacher distillation strategy, and multi-teacher feature fusion mechanism.

\subsection{Preliminaries}
\label{sec:Preliminaries}

In SSS, the training data typically comprises two distinct subsets: a small, meticulously annotated labeled dataset $D^l={(x_i^l, y_i^l)}_{i=1}^{B^l}$ and a large, unannotated unlabeled dataset $D^u={(x_i^u)}_{i=1}^{B^u}$. Here, $x_i$ represents the $i$-th input image, and $y_i$ is its corresponding pixel-wise ground truth mask. A fundamental characteristic of SSS tasks is the significant imbalance, where the size of $D^u$ is often orders of magnitude larger than $D^l$. The primary objective is to effectively harness the rich implicit information within this abundant unlabeled data to significantly boost model performance, transcending what can be achieved with labeled data alone.


FixMatch~\citep{fixmatch} serves as a robust baseline for SSS, ingeniously combining consistency regularization and pseudo-labeling within a mean-teacher framework. The training process involves iterating through mini-batches, each containing both labeled and unlabeled samples. For the labeled batch, images are fed into the student model $M_S$ and supervised directly using their ground-truth labels. The supervised loss for labeled samples is computed as:

\begin{equation}
    \mathcal{L}^l = \frac{1}{B^l}\sum_{i=1}^{B^l} CE(p_i^l, y_i^l),
\end{equation}
where $B^l$ denotes the batch size of labeled samples, $p_i^l$ represents the predicted segmentation map (probabilities) for the $i$-th labeled sample, $y_i^l$ is its corresponding ground-truth label, and $CE$ denotes the standard pixel-wise cross-entropy loss.

\definecolor{mygreen}{rgb}{0.0,0.5,0.0}
\renewcommand{\algorithmiccomment}[1]{\hfill\textcolor{mygreen}{\# #1}}

\begin{algorithm}[t]
\caption{Distillation and fusion process on unlabeled data in RS-MTDF}
\label{alg:forward}
\begin{algorithmic}[1]

\For{$x$ in loader\_u} 
    \newline \Comment{1. Strong augmentation for student}
    \State $x_s \gets \text{aug}_s(x)$ 
    \newline \Comment{2. Student encoder feature extraction}
    \State $f_S \gets E_S(x_s)$  
     \newline \Comment{3. Frozen VFM teacher feature extraction (DINO \& CLIP)}
    \State $f_T^{\text{dino}} \gets \text{DINO}(x_s)$
    \State $f_T^{\text{clip}} \gets \text{CLIP}(x_s)$  
     \newline \Comment{4. Translate student features to VFM feature spaces}
    \State $\hat{f}_S^{\text{dino}} \gets \text{MLP}_{\text{dino}}(f_S)$
    \State $\hat{f}_S^{\text{clip}} \gets \text{MLP}_{\text{clip}}(f_S)$ 
      \newline \Comment{5. Compute distillation losses}
    \State $\mathcal{L}_1 \gets \text{MSE}(\hat{f}_S^{\text{dino}}, f_T^{\text{dino}})$
    \State $\mathcal{L}_2 \gets \text{MSE}(\hat{f}_S^{\text{clip}}, f_T^{\text{clip}})$
    \State $\mathcal{L}_{\text{distill}} = (\mathcal{L}_1 + \mathcal{L}_2) / 2$
     \newline    \Comment{6. Multi-teacher feature fusion for decoder input}
    \State $\tilde{f}_S^{\text{dino}} \gets \text{Proj}_{\text{dino}}(\hat{f}_S^{\text{dino}})$ \Comment{Project DINO-aligned student features back to student's dimension}
    \State $\tilde{f}_S^{\text{clip}} \gets \text{Proj}_{\text{clip}}(\hat{f}_S^{\text{clip}})$ \Comment{Project CLIP-aligned student features back to student's dimension}

    \State $f_{\text{fused}} \gets \omega_S \cdot f_S + \omega_D \cdot (\tilde{f}_S^{\text{dino}} + \tilde{f}_S^{\text{clip}})$ \Comment{Weighted fusion of original and VFM-enhanced student features}

    \State $\text{features}[-1] \gets f_{\text{fused}}$ \Comment{Replace the highest-level student feature for decoder}
    \State $\text{pred} \gets D_S(\text{features})$ \Comment{Student decoder generates prediction}

 \Return $\text{pred}, \mathcal{L}_{\text{distill}}$

    \EndFor
\end{algorithmic}
\end{algorithm}




For the unlabeled batch, FixMatch leverages the teacher model $M_T$ to generate high-confidence pseudo-labels. The process adheres to the consistency regularization principle: an unlabeled image $x_i^u$ is first subjected to a weak augmentation strategy ($\text{aug}_w$) and passed through the teacher model to produce a prediction $p_i^w$. This prediction is then converted into a pseudo-label $\hat{y}_i^u$ by applying an \texttt{argmax} operation and a confidence threshold $\tau$. Subsequently, the same original unlabeled image $x_i^u$ is subjected to a strong augmentation strategy ($\text{aug}_s$) and fed into the student model to obtain prediction $p_i^s$. The student's prediction $p_i^s$ is then supervised by the pseudo-label $\hat{y}_i^u$ generated by the teacher from the weakly augmented view. This design enforces the assumption that the model's predictions should remain stable under varying input perturbations. The unsupervised loss for unlabeled samples is defined as:

\begin{equation}
    \mathcal{L}^u = \frac{1}{B^u} \sum_{i=1}^{B^u} \mathbb{1}(\max(p_i^w) \ge \tau) \cdot CE(p_i^s, \hat{y}_i^u),
\end{equation}
where $B^u$ is the batch size of unlabeled data, $\mathbb{1}(\cdot)$ is the indicator function that ensures only pseudo-labels with a maximum confidence score above the threshold $\tau$ are utilized, effectively filtering out low-confidence predictions.

Finally, the overall loss function, balancing supervised and unsupervised contributions, is expressed as:
\begin{equation}
    \mathcal{L} = \mathcal{L}^l + \alpha\mathcal{L}^u,
\end{equation}
where $\alpha$ is a hyperparameter balancing the importance of the unsupervised loss relative to the supervised loss.

\subsection{RS-MTDF}
\label{sec:3.2}


While FixMatch provides a solid foundation, its inherent coupling of labeled and unlabeled data learning processes can be suboptimal, especially when dealing with the significant distributional imbalance and domain shifts prevalent in remote sensing images. This often leads to the student model struggling to generalize effectively to the unlabeled set, as it might overfit to the limited labeled samples. Our core insight is that VFMs, pre-trained on vast and diverse datasets, possess remarkable generalization capabilities and rich semantic priors. We propose RS-MTDF to strategically leverage these VFMs as powerful external teachers within a distillation framework, guiding the student model to learn more generalizable and robust representations for remote sensing SSS. The detailed procedure of RS-MTDF is outlined in Algorithm~\ref{alg:forward}.




\begin{figure*}[t]
  \centering
   \includegraphics[width=\linewidth]{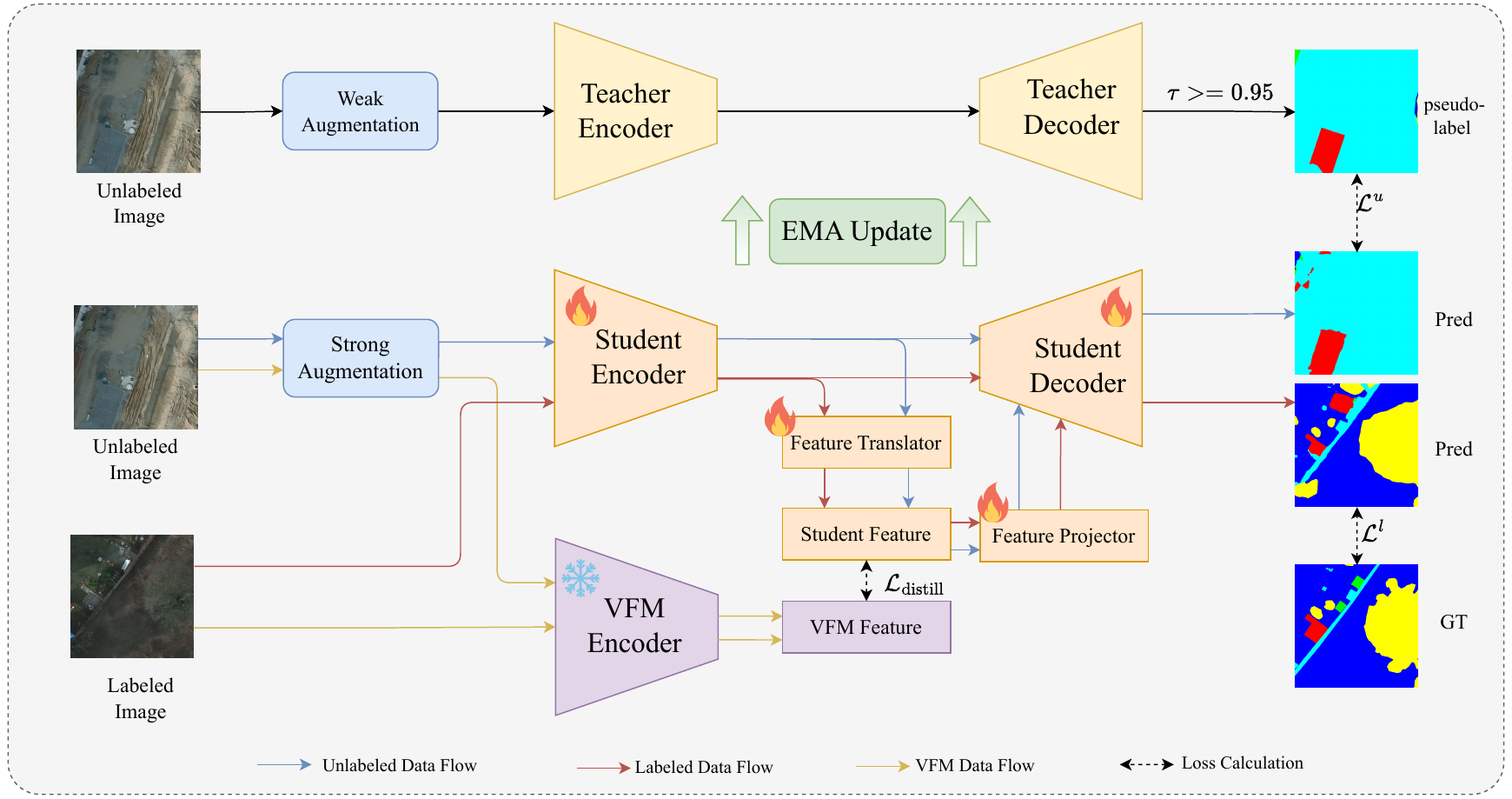}
   \caption{The overall architecture of our proposed RS-MTDF model. It consists of three main components: the student model (which learns from both labeled and unlabeled data), the teacher model (updated via EMA from the student for pseudo-labeling), and the frozen VFMs acting as auxiliary expert teachers. VFMs provide stable and rich prior knowledge through feature distillation and fusion.}
   \label{fig:framework}
\end{figure*}

\subsubsection{Archetecture Overview}

The overall architecture of our proposed RS-MTDF framework is illustrated in Figure~\ref{fig:framework}. RS-MTDF retains the fundamental structure of FixMatch by adopting a mean teacher framework, where the student model $M_S$ and teacher model $M_T$ (which share the same architectural backbone) are intertwined. The teacher model's parameters $\theta^{\text{teacher}}$ are updated as an exponential moving average (EMA) of the student model's parameters $\theta^{\text{student}}$:
\begin{equation}
\theta_t^{\text{teacher}} = \alpha \cdot \theta_{t-1}^{\text{teacher}} + (1 - \alpha) \cdot \theta_t^{\text{student}}
\end{equation}
where $\alpha$ is the momentum coefficient, typically set close to 1. This EMA update provides a more stable target for the student model's learning.

The core of RS-MTDF's innovation lies in introducing multiple frozen VFMs as auxiliary, powerful external teachers: DINOv2 \citep{dinov2} for its strong spatial awareness and CLIP \citep{clip} for its broad semantic understanding. These VFMs are kept fixed throughout training, providing stable and rich knowledge sources. During training, both labeled and unlabeled images are fed through the student encoder and the frozen VFM teacher encoders to extract their respective feature representations.

RS-MTDF integrates VFM knowledge into the student's learning process through two main stages:
\begin{enumerate}
    \item VFM-Guided Feature Distillation: We perform feature-level distillation at the encoder stage. This involves projecting student features into the VFM feature spaces via dedicated feature translators and minimizing a distillation loss to align student features with the robust representations from DINOv2 and CLIP. This process enhances the student encoder's generalization capabilities.
    \item Multi-Teacher Feature Fusion: To further exploit the distilled knowledge, the VFM-aligned student features are then projected back to the student's dimension and fused with the original student's high-level features. This fused representation replaces the standard highest-level encoder output, enriching the input to the decoder and ensuring VFM guidance influences the final segmentation prediction.
\end{enumerate}
This comprehensive integration ensures that the student model benefits from the VFMs' strong priors, enabling it to better handle the distribution mismatch and complex patterns in remote sensing images.




\subsubsection{VFM-Guided Feature Distillation}


Due to inherent architectural differences between the student model and the VFM teacher models, their extracted feature representations often have mismatched dimensions. To bridge this gap, we introduce dedicated feature translator modules that project the student features into the corresponding VFM teacher feature spaces.

Specifically, we utilize a dedicated feature translator module to bridge this representational gap. Both the student encoder and VFM teachers are based on Transformer architectures. We extract their output representations from the final encoder layer. The student's feature map is denoted as $f_S \in \mathbb{R}^{B \times N \times d_s}$, and a generic VFM teacher's feature map is $f_T \in \mathbb{R}^{B \times N \times d_t}$. Here, $B$ is the batch size, $N$ is the number of patch tokens (or flattened spatial locations), and $d_s$, $d_t$ are the embedding dimensions of the student and teacher, respectively.

Each VFM teacher is associated with its own distinct feature translator module. This translator is implemented as a simple, non-linear two-layer Multilayer Perceptron (MLP) to project the student features $f_S$ into the respective VFM feature spaces. For a given teacher $t \in \{\text{DINOv2, CLIP}\}$, its translator $\text{MLP}_t$ is defined as:
\begin{equation}
\hat{f}_S^{(t)} = \text{MLP}_t(f_S) = W_2 \cdot \text{ReLU}(W_1 \cdot f_S + b_1) + b_2,
\end{equation}
where $W_1, W_2, b_1, b_2$ are parameters of the MLP.




Following this feature translation, we enforce alignment between the projected student features $\hat{f}_S^{(t)}$ and the frozen VFM teacher features $f_T^{(t)}$ using a Mean Squared Error (MSE) loss. This distillation objective compels the student encoder to mimic the high-quality feature distributions and robust semantic representations learned by the VFMs, thereby imparting their generalization capabilities. The distillation loss for the $t$-th teacher is calculated as:
\begin{equation}
\mathcal{L}_{\text{distill}}^{(t)} = \frac{1}{BN} \sum_{i=1}^{B} \sum_{j=1}^{N} \left\| \hat{f}_S^{(t)}[i,j] - f_T^{(t)}[i,j] \right\|_2^2,
\end{equation}
where $f_S^{(t)}[i,j]$ and $f_T^{(t)}[i,j]$ refer to the $j$-th patch feature of the $i$-th sample from the projected student and VFM teacher, respectively. The total distillation loss $\mathcal{L}_{\text{distill}}$ used in the overall objective is the average of individual distillation losses from both VFM teachers: 

\begin{equation}
\mathcal{L}_{\text{distill}} = \frac{1}{2} (\mathcal{L}_{\text{distill}}^{\text{DINOv2}} + \mathcal{L}_{\text{distill}}^{\text{CLIP}}).
\end{equation}

Finally, the overall loss function of RS-MTDF combines the supervised loss $\mathcal{L}^l$, the unsupervised consistency loss $\mathcal{L}^u$, and the total distillation loss $\mathcal{L}_{\text{distill}}$:
\begin{equation}
\mathcal{L} = \lambda_l \mathcal{L}^l + \lambda_u \mathcal{L}^u + \lambda_d \mathcal{L}_{\text{distill}}
\end{equation}
where $\lambda_l$, $\lambda_u$, and $\lambda_d$ are hyperparameters that balance the contributions of supervised learning, unsupervised consistency regularization, and the VFM-guided distillation, respectively.

\subsubsection{Multi-Teacher Feature Fusion}

To maximize the benefits of the rich semantic and spatial knowledge transferred from the multi-teacher VFMs, RS-MTDF integrates the VFM-aligned student features directly into the student's decoding pathway. This multi-teacher feature fusion mechanism ensures that the distilled knowledge not only influences early encoder learning but also explicitly refines the final pixel-wise predictions.

Specifically, the projected student features $\hat{f}_S^{\text{DINOv2}}$ (ali-gned with DINOv2's space) and $\hat{f}_S^{\text{CLIP}}$ (aligned with CLIP's space) are first individually projected back to the original student feature dimension using separate learnable linear layers, $\text{Proj}_{\text{DINOv2}}$ and $\text{Proj}_{\text{CLIP}}$ respectively. This re-projection ensures dimensional compatibility for subsequent fusion. These re-projected, VFM-enhanced features, which now carry the refined semantic and spatial information from their respective VFM teachers, are then combined with the original high-level student encoder feature $f_S$ to form a comprehensive, knowledge-enriched representation, $f_{\text{fused}}$:


\begin{equation}
\begin{split}
f_{\text{fused}} &= \omega_S \cdot f_S + \\
& \quad \omega_D \cdot \left( \text{Proj}_{\text{DINO}}(\hat{f}_S^{\text{DINO}}) + \text{Proj}_{\text{CLIP}}(\hat{f}_S^{\text{CLIP}}) \right)
\end{split}
\end{equation}

This weighted fusion strategy utilizes fusion weights $\omega_S$ and $\omega_D$. These weights are determined empirically to ensure that the student's own learned representations retain primary influence, while judiciously incorporating the powerful and generalized priors from the VFM teachers




The $f_{\text{fused}}$ feature explicitly replaces the original highest-level feature output from the student encoder before it is passed to the decoder for final pixel-wise prediction. For the decoder, we adopt a hierarchical design inspired by DPTHead \citep{ranftl2021vision}. This decoder progressively refines multi-scale features extracted from various layers of the student encoder. We utilize four intermediate features from the student encoder, which are first projected to a common channel dimension via $1 \times 1$ convolutions and then resized to a unified spatial resolution using transposed convolutions. These unified features, including the $f_{\text{fused}}$ at the highest level, are then processed through a series of feature fusion blocks to enable robust inter-level refinement and rich context integration. The final segmentation map is generated via a convolutional prediction head and subsequently upsampled to the original input resolution. This sophisticated decoder structure, combined with the VFM-guided $f_{\text{fused}}$, allows for effective integration of broad semantic information across different scales and spatial resolutions, ensuring that the final predictions benefit from both the student's learned representations and the powerful, generalized priors from the VFM teachers.

\section{Experiment}
\label{sec:exp}

This section presents a comprehensive evaluation of the proposed RS-MTDF framework. We detail our experimental setup, compare RS-MTDF with state-of-the-art SSS methods on three widely-used remote sensing benchmarks, and conduct extensive ablation studies to validate the contribution of each proposed module.

\subsection{Experiment Setup}

\subsubsection{Dataset}
To assess the performance and generalization capabilities of RS-MTDF, we conduct experiments on three diverse and challenging remote sensing semantic segmentation datasets:


\begin{enumerate}
    \item \textbf{ISPRS Potsdam}~\citep{potsdam}: This dataset was proposed to promote high-resolution remote sensing image segmentation. It consists of 38 large-scale aerial images from Germany, each with a resolution of 0.05m/pixel and a size of $6000 \times 6000$ pixels. The semantic categories include \textit{Building}, \textit{Low Vegetation}, \textit{Tree}, \textit{Car}, and \textit{Background}.

    \item \textbf{LoveDA}~\citep{loveda}: This dataset contains 5,987 images captured from three cities in China. Each image has a resolution of 0.3m/pixel and a size of $1024 \times 1024$ pixels. The dataset is annotated with seven categories: \textit{Background}, \textit{Building}, \textit{Road}, \textit{Water}, \textit{Barren}, \textit{Forest}, and \textit{Agriculture}.

    \item \textbf{DeepGlobe Land Cover}~\citep{deepglobe}: This dataset includes satellite images from Thailand, Indonesia, and India, with a resolution of 0.5m/pixel and an image size of $2448 \times 2448$. The labeled categories are \textit{Urban}, \textit{Agriculture}, \textit{Rangeland}, \textit{Forest}, \textit{Water}, and \textit{Barren}.
\end{enumerate}


Following the common experimental protocol established in \citep{dwl}, all images are uniformly cropped to $512 \times 512$ patches. Each dataset is then partitioned into training, validation, and test sets with a ratio of 6:2:2. For the training set, to simulate low-label scenarios characteristic of SSS, we further divide the data into labeled and unlabeled subsets using three different label proportions: 1\%, 5\%, and 10\%.

\subsubsection{Implementation Details}

We implement our method using PyTorch. During training, we adopt DINOv2 small~\citep{dinov2} as the encoder and use a DPT~\citep{Ranftl2020} head as the decoder.
During training, we utilize the AdamW optimizer \citep{adamw} with a learning rate of $5 \times 10^{-6}$ for the encoder and $2 \times 10^{-4}$ for the decoder. The optimizer parameters $\beta$ are set to $(0.9, 0.999)$, and a weight decay of 0.01 is applied. The batch size for training is set to 8.
The confidence threshold $\tau$ for filtering pseudo-labels in consistency regularization is fixed at 0.95. All loss weights (\textit{i.e.}, $\lambda_l$ for supervised loss, $\lambda_u$ for unsupervised loss, and $\lambda_d$ for distillation loss) are set to $\frac{1}{3}$, ensuring a balanced contribution from each component.
For data augmentation, we apply standard settings. Weak augmentations include simple resizing, random cropping, and horizontal flipping. Strong augmentations, used for the student model's unlabeled input, further incorporate color jitter and CutMix \citep{cutmix} to increase perturbation diversity and enhance robustness.
For VFM teacher guidance, we employ DINOv2-Base (for spatial priors) and CLIP ViT-L/14 \citep{clip} (for broad semantic understanding) as our frozen teacher models. These models are chosen for their complementary strengths and robust pre-trained representations. All experiments are conducted for 60 epochs on NVIDIA A800 GPUs.

\subsubsection{Evaluation MMetrics}

To comprehensively evaluate the segmentation performance, we employ three widely recognized metrics: Intersection-over-Union (\textit{IoU}), F$_1$-score, and the Kappa coefficient. These metrics are computed per-class and then averaged (for mIoU, mF$_1$) to reduce the impact of class imbalance. They are formally defined as follows:

\begin{equation}
\textit{IoU} = \frac{TP}{TP + FP + FN},
\end{equation}

\begin{equation}
F_1 = \frac{2TP}{2TP + FP + FN},
\end{equation}



\begin{equation}
\textit{Kappa} = \frac{OA - PRE}{1 - PRE},
\end{equation}

where TP, TN, FP, and FN denote the number of true positives, true negatives, false positives, and false negatives, respectively.

\subsection{Comparison with SOTA Methods}
\subsubsection{Comparison Methods}
Our RS-MTDF method is benchmarked against several state-of-the-art SSS approaches, including general SSS methods adapted to remote sensing and remote sensing-specific SSS methods:
\begin{itemize}
    \item \textbf{FullySup}: Represents the upper bound performance, trained using 100\% of the labeled training data as an oracle model.
    \item \textbf{CutMix}~\citep{cutmix}: A strong data augmentation technique where parts of two unlabeled images are mixed to create new training samples.
    \item \textbf{CCT}~\citep{CCT}: Enforces consistent predictions across multiple perturbed models or views of the same input.
    \item \textbf{CPS}~\citep{chen2021-CPS}: Utilizes predictions from two different models as pseudo-labels to supervise each other, promoting robust learning.
    \item \textbf{LSST}~\citep{LSST}: An adaptive thresholding method that assigns pseudo-labels by linearly sampling confidence thresholds for different classes.
    \item \textbf{FixMatch}~\citep{fixmatch}: A foundational self-training method based on thresholding pseudo-labels from weakly augmented inputs to supervise strongly augmented inputs.
    \item \textbf{AllSpark}~\citep{allsparkrebornlabeledfeatures}: A Transformer-based method that employs a channel-wise cross-attention mechanism to leverage labeled features for unlabeled data, aiming to alleviate distribution shifts.
    \item \textbf{UniMatchv2}~\citep{unimatchv2}: An advanced FixMatch variant that explores a broader space of perturbations and diverse augmentation strategies, achieving strong performance in semi-supervised semantic segmentation on natural images.
    \item \textbf{DWL}~\citep{dwl}: A framework designed for semi-supervised segmentation that decouples learning processes and addresses issues like incorrect pseudo-label propagation and long-tailed class distributions in remote sensing.
    \item \textbf{MUCA}~\citep{semisupervisedsemanticsegmentationremote}: A method for remote sensing SSS that integrates multi-scale uncertainty consistency regularization and a cross-teacher-student attention mechanism to handle complex remote sensing image characteristics.
\end{itemize}

\subsubsection{Main Results}
\begin{table*}[!t]
\caption{Comparison results with SOTA methods on ISPRS Potsdam dataset.\ The best results are highlighted in bold. IoU, mIoU, mF1, and Kappa are represented as percentages.\label{tab:table5}}
\centering
\normalsize
\scalebox{0.80}{ 
\begin{tabular*}{\textwidth}{@{\extracolsep{\fill}} c c c @{\hspace{0.4em}} c @{\hspace{0.4em}} c @{\hspace{0.4em}} c @{\hspace{0.4em}} c @{\hspace{0.4em}} c @{}} 
		\toprule
		\textbf{Ratio} & \textbf{Model} &\multicolumn{5}{c}{\textbf{IoU}}  & \textbf{mIoU}/\textbf{mF1}/\textbf{Kappa} \\
		\cline{3-7}
		~ & ~ & Building & \makecell{Low \\ vegetation}  & Tree & Car & \makecell{Back \\ ground} &  ~\\
		\midrule
		\multirow{11}{*}{\textbf{1\%}} & CutMix~\citep{cutmix} &  55.58 & 42.05 &49.72&50.86&39.40&47.52 / 64.21 / 0.5121\\
		~& CCT~\citep{CCT} &  54.48 & 61.28 &48.56&52.95&60.71&55.59 / 71.34 / 0.5761\\
        ~& CPS~\citep{chen2021-CPS} &  59.35 & 69.16 &62.89&59.88&66.33&63.52 / 77.63 / 0.6539\\
		~& LSST\citep{LSST} & 68.74 & 75.24 &54.74&62.09&68.80&65.92 / 79.25 / 0.6847\\
		~& FixMatch~\citep{fixmatch} & 76.95 & 71.59 &64.71&65.85&72.81&70.38 / 82.53 / 0.7287\\
		~& UniMatch~\citep{unimatch} & 76.52 & 70.99 &65.44&66.62&72.64&70.44 / 82.59 / 0.7291\\
        ~& DWL\citep{dwl} & 72.34 & \textbf{77.08} &62.74&62.57&72.22&69.39 / 81.79 / 0.7191\\
		~& AllSpark~\citep{allsparkrebornlabeledfeatures} & 83.70 & 65.92 &59.64&69.77&75.31&70.87 / 82.68 / 0.7827\\
		~& MUCA~\citep{semisupervisedsemanticsegmentationremote} & 84.56 & 66.98 &56.96&71.52&\textbf{76.}\textbf{64}&71.33 / 82.92 / 0.7880\\
         ~&Unimatchv2 ~\citep{unimatchv2} & 91.43 & 75.17 & 73.93 & 82.50 & 53.48 & 75.30 / 82.10 / 0.7904 \\
            ~& Our & \textbf{91.59} & 75.21 &\textbf{ 74.93} & \textbf{85.31} & 57.91 & \textbf{76.99 / 83.22 / 0.8044 }\\
		\midrule
		\multirow{11}{*}{\textbf{5\%}}& CutMix~\citep{cutmix} & 52.94 & 68.86 &41.51&58.33&54.82&55.29 / 70.79 / 0.5783\\
		~& CCT~\citep{CCT} & 72.90 & 80.25 &64.23&58.32&74.42&70.02 / 82.12 / 0.7236\\
        ~& CPS~\citep{chen2021-CPS} & 76.53 & 84.34 &57.98&69.45&75.39&72.74 / 83.78 / 0.7492\\
		~& LSST~\citep{LSST} & 69.26 & 84.55 &67.33&67.49&73.86&72.50 / 83.67 / 0.7399\\
        ~& FixMatch~\citep{fixmatch} & 78.12 & 74.87 &68.89&66.58&75.30&72.75 / 84.15 / 0.7497\\
		~& UniMatch~\citep{unimatch} & 78.24 & 73.59 &67.17&66.64&75.07&72.14 / 83.73 / 0.7432\\
        ~& DWL~\citep{dwl} & 74.81 & 85.64 &66.38&62.99&75.68&73.10 / 84.22 / 0.7507\\
		~& AllSpark~\citep{allsparkrebornlabeledfeatures} & 85.57 & 67.62 &60.61&73.48&77.15&72.88 / 84.04 / 0.7989\\
		~& MUCA~\citep{semisupervisedsemanticsegmentationremote} & 88.45 & 69.53 &61.39&74.18&\textbf{79.56}&74.62 / 85.15 / 0.8166\\
             ~&Unimatchv2 ~\citep{unimatchv2} & 93.25 & 76.61& 74.79 & 85.61 & 69.46 & 79.94 / 85.58 / 0.8250 \\
        ~& Our & \textbf{92.92} & \textbf{76.89} & \textbf{74.97} &\textbf{ 88.34} & 74.65 & \textbf{81.55 / 86.15 / 0.8324 }\\
		\midrule
		\multirow{11}{*}{\textbf{10\%}}& CutMix~\citep{cutmix} & 64.55 & 80.99 &64.79&65.50&68.01&68.77 / 81.34 / 0.7109\\
		~& CCT~\citep{CCT} & 73.09 & 83.94 &61.12&60.45&73.06&70.33 / 82.27 / 0.7265\\
        ~& CPS~\citep{chen2021-CPS} & 77.80 & 87.15 &61.12&68.48&75.89&74.09 / 84.55 / 0.7533\\
		~& LSST~\citep{LSST} & 70.92 & 86.06 &68.91&70.22&74.89&74.20 / 84.95 / 0.7549\\
        ~& FixMatch~\citep{fixmatch} & 77.97 & 76.17 &70.09&70.97&76.14&74.27 / 85.20 / 0.7606\\
		~& UniMatch~\citep{unimatch} & 77.34 & 87.75 &70.79&56.65&76.46&73.80 / 84.52 / 0.7599\\
        ~& DWL~\citep{dwl} & 76.37 & 88.42 &66.54&64.37&77.14&74.57 / 85.16 / 0.7628\\
		~& AllSpark~\citep{allsparkrebornlabeledfeatures} & 86.29 & 69.83 &64.17&75.23&78.31&74.76 / 85.35 / 0.8144\\
		~& MUCA~\citep{semisupervisedsemanticsegmentationremote} & 88.02 & 70.58 &64.53&75.20&\textbf{79.92}&75.65 / 85.90 / 0.8245\\
                     ~&Unimatchv2 ~\citep{unimatchv2} & 93.43&\textbf{78.01}&\textbf{75.66}&86.93&76.06& \textbf{82.02} / \textbf{86.86} / \textbf{0.8411} \\
        ~& Our & \textbf{93.63} & 77.23 & 75.32 & \textbf{87.44} & 74.50 & 81.62 / 86.64 / 0.8362 \\
		\midrule
		\textbf{100\%} & FullySup &94.74 & 80.38 & 78.02 & 87.33 & 80.08 &  84.11 / 88.61 / 0.8645 \\
		\bottomrule
        
	\end{tabular*}
    }
     \label{tab:potsdam}
\end{table*}
\begin{table*}[!t]
\caption{Comparison results with other SOTA methods on the LoveDA dataset. The best results are in bold. IoU, mIoU, mF1, and Kappa1 are represented as percentages. \label{tab:table6}}
\centering
\small
\normalsize
\resizebox{\textwidth}{!}{
\scalebox{0.1}{ 
\begin{tabular*}{\textwidth}{@{\extracolsep{\fill}} c c c @{\hspace{0.5em}} c @{\hspace{0.5em}} c @{\hspace{0.5em}} c @{\hspace{0.5em}} c @{\hspace{0.5em}} c @{\hspace{0.5em}} c @{\hspace{0.5em}} c @{}}
		\cline{1-10}
		\textbf{Ratio} & \textbf{Model} &\multicolumn{7}{c}{\textbf{IoU}}  & \textbf{mIoU}/\textbf{mF1}/\textbf{Kappa} \\
		\cline{3-9}
		~ & ~ & \makecell{Back \\ ground} &  Building &Road & Water & Barren & Forest & \makecell{Agri- \\ culture} &  ~\\
		\cline{1-10}
	\multirow{11}{*}{\textbf{1\%}}
		~& CutMix~\citep{cutmix} & 36.04 & 24.69 &10.03&24.60&3.43&6.67&10.19&16.52 / 26.85 / 0.1362\\
		~& CCT ~\citep{CCT} & 37.16 & 22.41 &27.86&43.98&14.51&25.38&36.67&29.71 / 44.99 / 0.3517\\
        ~& CPS~\citep{chen2021-CPS} & 46.52 & 20.87 &27.85&50.55&0.01&33.16&34.60&30.51 / 44.28 / 0.3802\\
		~& LSST~~\citep{LSST}  & 44.73 & 41.90 &39.90&62.65&29.27&31.26&48.29&42.57 / 59.00 / 0.4817\\
        ~& FixMatch~\citep{fixmatch} & 46.78 & 51.20 &50.21&67.27&11.53&\textbf{36.79}&50.26&44.86 / 60.02 / 0.5175\\
		~& UniMatch~\citep{unimatch} & 46.53 & 51.38 &49.36&\textbf{67.74}&10.86&33.40&52.28&44.51 / 59.51 / 0.5175\\
		~& DWL~\citep{dwl} & 48.74 & 56.79 &51.59&63.42&22.56&35.20&55.38&47.67 / 63.40 / 0.5534\\
        ~& AllSpark~\citep{allsparkrebornlabeledfeatures} & 63.87 & 47.70 &46.05&61.52&35.31&30.94&55.64&48.72 / 63.29 / 0.5502\\
		~& MUCA~\citep{semisupervisedsemanticsegmentationremote} & \textbf{64.89} & 56.03 &47.14&63.86&\textbf{35.81}&22.57&\textbf{58.18}&49.78 / 63.48 / 0.5753\\
        &Unimatchv2~\citep{unimatchv2}& 52.91 & 59.97 & 52.28 & 65.54 & 24.59 & 37.74 & 60.38 & 50.49 / 64.87 / 0.6051 \\
        & Our & 52.33 & \textbf{59.98} & \textbf{55.48} & 64.56 & 32.52 & 36.51 & 57.07 &\textbf{51.21 / 65.93 /  0.5968}\\
			\cline{1-10}
	\multirow{11}{*}{\textbf{5\%}}& CutMix~\citep{cutmix} & 41.48 & 41.62 &38.77&47.44&14.69&28.09&31.05&34.73 / 50.65 / 0.3692\\
		~& CCT~\citep{CCT} & 46.80 & 44.62 &46.80&60.95&24.83&29.03&44.30&42.48 / 58.74 / 0.4850\\
        ~& CPS~\citep{chen2021-CPS} & 48.90 & 49.64 &47.97&60.27&4.67&36.09&47.32&42.12 / 56.90 / 0.4976\\
		~& LSST~\citep{LSST}& 51.48 & 45.66 &52.66&67.63&33.52&35.80&48.60&47.91 / 64.10 / 0.5434\\
        ~& FixMatch~\citep{fixmatch}& 45.40 & 53.05 &51.22&66.73&28.53&27.25&54.30&44.64 / 62.45 / 0.5378\\
		~& UniMatch~\citep{unimatch} & 50.20 & 54.49 &50.46&67.18&26.79&30.06&54.86&47.72 / 63.46 / 0.5543\\
        ~& DWL~\citep{dwl} & 48.75 & 55.00 &51.53&69.49&29.46&36.59&52.11&48.99 / 64.88 / 0.5597\\
		~& AllSpark~\citep{allsparkrebornlabeledfeatures} & 65.09 & 55.06 &47.59&67.10&34.67&26.86&51.87&49.75 / 64.91 / 0.5682\\
		~& MUCA ~\citep{semisupervisedsemanticsegmentationremote}& \textbf{67.29} & 56.04 &48.37&61.02&36.21&30.76&57.09&50.97 / 64.92 / 0.5856\\
         &Unimatchv2~\citep{unimatchv2}&53.82 & 64.49 & 51.80 & 69.81 & 31.35 & 39.62 & 61.61 & 53.21 / 67.30 / 0.6208 \\
        & Our & 55.49 & \textbf{64.10} & \textbf{55.96} & \textbf{70.15} & 33.67 & \textbf{41.07} & \textbf{63.81}& \textbf{54.89 / 68.83 / 0.6396} \\
			\cline{1-10}
		\multirow{11}{*}{\textbf{10\%}}&  CutMix~\citep{cutmix} & 46.73 & 49.60 &47.36&59.99&29.06&37.77&40.60&44.44 / 60.99 / 0.4837\\
		~& CCT~\citep{CCT} & 44.07 & 45.22 &47.65&57.12&24.41&32.50&45.07&42.29 / 58.73 / 0.4762\\
        ~& CPS~\citep{chen2021-CPS} & 51.30 & 54.93 &52.57&53.37&18.39&37.59&53.24&45.91 / 61.78 / 0.5479\\
		~& LSST~\citep{LSST} & 50.69 & 49.50 &52.63&69.85&27.25&36.24&52.06&48.32 / 64.17 / 0.5565\\
        ~& FixMatch~\citep{fixmatch} & 52.02 &  55.59 &53.20& 57.91& 25.86&40.83& 57.50& 48.99 / 64.97 / 0.5676\\
		~& UniMatch~\citep{unimatch} & 51.80 & 53.95 & 51.17&58.15&25.60&38.72&54.86&47.75 / 63.86 / 0.5639\\
        ~& DWL~\citep{dwl} & 49.94 &  56.66 & 53.89&70.35& 30.62&41.49&53.13&50.87 / 66.64 / 0.5753\\
		~& AllSpark~\citep{allsparkrebornlabeledfeatures} & 67.13 & 56.16 &40.67&63.58&32.54&32.03&56.91&49.86 / 63.97 / 0.5751\\
		~& MUCA~\citep{semisupervisedsemanticsegmentationremote} & \textbf{68.69} & 58.20 &41.82&65.62&\textbf{37.09}&35.01&57.38&51.97 / 66.72 / 0.5901\\
         &Unimatchv2~\citep{unimatchv2}& 50.10 & 60.26 & 57.89 & 68.21 & 32.44 & 43.81 & 60.62 & 53.33 / 67.65 / 0.6137 \\
        & Our &54.24 & \textbf{64.54} &\textbf{ 55.93} & \textbf{70.79} & 32.48 & \textbf{42.71} & \textbf{62.64} & \textbf{54.76 / 68.65 / 0.6306} \\
		\cline{1-10}
		\textbf{100\%} & FullySup & 68.84 & 58.57 &48.02&70.39&43.28&38.59&62.30&55.71 / 69.10 / 0.6291\\
	\cline{1-10}
	\end{tabular*}
    }
    }
    \label{tab:loveda}
\end{table*}

We conduct extensive experiments across the three remote sensing benchmarks: ISPRS Potsdam, LoveDA, and DeepGlobe Land Cover. The detailed quantitative comparison results, including per-class IoU, mean IoU (mIoU), mean F$_1$-score (mF$_1$), and Kappa coefficient, are presented in Table~\ref{tab:potsdam} (ISPRS Potsdam), Table~\ref{tab:loveda} (LoveDA), and Table~\ref{tab:deepglobe} (DeepGlobe), respectively.

Experimental results show that our method consistently outperforms existing state-of-the-art approaches across the ISPRS Potsdam, LoveDA, and DeepGlobe datasets. Notably, the improvements are achieved under extremely low label ratios of 1\%, 5\%, and 10\%, respectively. These results strongly demonstrate the effectiveness of our approach in low-label regimes and its potential for practical deployment in real-world remote sensing scenarios. 

Overall, our method consistently demonstrates competitive or superior performance compared to existing state-of-the-art approaches across all three datasets, particularly under challenging low-label ratios (1\%, 5\%, and 10\%). These results strongly underscore the effectiveness of RS-MTDF in leveraging VFM priors for robust semi-supervised learning in data-scarce scenarios, highlighting its potential for practical deployment in real-world remote sensing applications.

On the ISPRS Potsdam dataset (Table~\ref{tab:potsdam}), RS-MTDF showcases excellent performance. While UniMatchv2 slightly outperforms our method at the 10\% label ratio (82.02\% mIoU \textit{vs.} 81.62\%), RS-MTDF achieves the best results under the more challenging 1\% (76.99\% mIoU) and 5\% (81.55\% mIoU) label settings. This indicates RS-MTDF's superior ability to generalize with minimal annotations. From a per-class perspective, our method achieves remarkably high IoU scores on small-scale object categories such as \textit{Car} (85.31\% at 1\%, 88.34\% at 5\%, 87.44\% at 10\%) and \textit{Building} (consistently highest or second-highest), demonstrating its efficacy in alleviating the multi-scale challenge inherent in remote sensing segmentation. Conversely, performance on the \textit{Background} class is relatively suboptimal. This could be attributed to the strong emphasis on foreground objects induced by our feature-level distillation process, which might inadvertently suppress less discriminative background features. Furthermore, the inherent semantic ambiguity of background regions in complex urban scenes can further hinder accurate segmentation.

On the LoveDA dataset (Table~\ref{tab:loveda}), RS-MTDF consistently achieves competitive or superior performance across all supervision ratios. Notably, under the extremely low-label 1\% setting, our model achieves a mIoU of 51.21\%, outperforming UniMatchv2 (50.49\%) and MUCA (49.78\%). Under the 5\% and 10\% label ratios, RS-MTDF continues to yield strong results with mIoU scores of 54.89\% and 54.76\% respectively, closely approaching the fully supervised upper bound (55.71\%). Our method surpasses UniMatchv2 on both settings, demonstrating improved scalability and generalization across varying supervision levels. These results confirm RS-MTDF's capability to effectively leverage limited annotations for strong segmentation performance in complex urban and rural scenes. While RS-MTDF may not always be the absolute best in every single class (\textit{e.g.}, \textit{Road} under 10\%), it demonstrates a balanced and robust performance across all categories, indicating its strong generalization ability without overfitting to dominant classes.

On the DeepGlobe Land Cover dataset (Table~\ref{tab:deepglobe}), RS-MTDF achieves either the best or second-best performance across most semantic categories. Significantly, it obtains the highest mIoU scores under both the 1\% (67.15\%) and 10\% (73.48\%) label ratios, showcasing its effectiveness and scalability across varying supervision levels and diverse land cover types.

\begin{table*}[!t]
\caption{Comparison results with SOTA methods on DeepGlobe dataset.\ The best results are highlighted in bold. IoU, mIoU, mF1, and Kappa are represented as percentages.\label{tab:table5}}
\centering
\normalsize
\scalebox{0.80}{ 
\begin{tabular*}{\textwidth}{@{\extracolsep{\fill}} c c c @{\hspace{0.5em}} c @{\hspace{0.5em}} c @{\hspace{0.5em}} c @{\hspace{0.5em}} c @{\hspace{0.5em}} c @{\hspace{0.5em}} c @{}} 
		\toprule
		\textbf{Ratio} & \textbf{Model} &\multicolumn{6}{c}{\textbf{IoU}}  & \textbf{mIoU}/\textbf{mF1}/\textbf{Kappa} \\
		\cline{3-8}
		~ & ~ &Urban &  \makecell{Agri- \\ culture} & \makecell{Range- \\ land}  &Forest & Water&  Barren&~\\
		\midrule
		\multirow{11}{*}{\textbf{1\%}} & CutMix~\citep{cutmix} &  50.44 & 58.97 & 6.61 &35.84&19.78 &  1.42 &  28.84/ 40.38 / 0.4024\\
		~& CCT~\citep{CCT} & 70.86 & 70.64 &  11.03 & 62.44 &  28.76 &  27.66 &  45.23 / 58.41 / 0.6039 \\
        ~& CPS~\citep{chen2021-CPS} &  80.94 &70.66 &  1.16 & 63.91 & 27.45 &  0.79 &  40.82 / 49.53 / 0.5945 \\
		~& LSST\citep{LSST} & 79.35 &  73.41 &  21.60 &  60.76 &  30.40 & 25.64 &  48.53 / 61.95 / 0.6314 \\
		~& FixMatch~\citep{fixmatch} & 82.51 &  74.10 &  18.79 & 67.65 & 44.72 &  32.74 &  53.42 / 66.50 / 0.6591 \\
		~& UniMatch~\citep{unimatch} & 80.54 & 70.72 & 20.71 & 65.48 &  34.09 &  9.24 & 46.80 / 58.88 / 0.6114 \\
        ~& DWL\citep{dwl} &  81.66 & 75.40 &  21.82 &  67.10 & 63.04 & 35.27 &  57.38 / 70.25 / 0.6758 \\
         ~&Unimatchv2 ~\citep{unimatchv2} & 84.57 & 83.31 & 25.01 & \textbf{72.38} & \textbf{74.38} & 58.43 & 66.35 / 77.59 / 0.7833 \\
            ~& Our & \textbf{86.59} & \textbf{84.33} & \textbf{29.0} & 70.26 & 73.67 & \textbf{59.04}& \textbf{67.15 / 78.47 / 0.7881} \\
		\midrule
		\multirow{11}{*}{\textbf{5\%}}& CutMix~\citep{cutmix} & 81.07 &  74.20 &  6.78 & 64.55 &  55.88 &  33.06 &  52.59 / 64.55 / 0.6540 \\
		~& CCT~\citep{CCT} & 81.20 &  76.14 &  12.38 & 64.05 & 49.88 & 42.97 & 54.44 / 67.14 / 0.6791\\
        ~& CPS~\citep{chen2021-CPS} &  84.15 & 78.67 & 11.31 &71.15 & 57.49 & 43.23 & 57.67 / 69.38 / 0.7183 \\
		~& LSST\citep{LSST} & 84.26 & 81.67 & 30.71 & 68.25 & 65.62 & 55.16 & 64.28 / 76.24 / 0.7528\\
		~& FixMatch~\citep{fixmatch} & 85.31 & 82.96 & 32.22 & 67.47 & 69.76 & 59.09 &  66.13 / 78.09 / 0.7687 \\
		~& UniMatch~\citep{unimatch} & 84.13 &  81.36 & 30.69 & 69.83 & 65.84 & 54.38 & 64.37 / 76.69 / 0.7535 \\
        ~& DWL\citep{dwl} & 86.08&  83.43&  36.62 & 70.22 &  70.77 & 59.86 &  67.83 / 79.56 / 0.7788 \\
         ~&Unimatchv2 ~\citep{unimatchv2} & 86.25 & 85.84 & \textbf{41.22} & \textbf{76.14} & 80.94 & 60.58 & \textbf{71.83 / 82.44 / 0.8164} \\
            ~& Our &\textbf{ 86.79} &\textbf{ 86.00 }& 39.13 & 76.01 & \textbf{80.99} & \textbf{61.27 }& 71.70 / 82.24 / 0.8164 \\
		\midrule
		\multirow{11}{*}{\textbf{10\%}}& CutMix~\citep{cutmix} &  63.25 & 73.44 &  26.59 & 64.42 & 60.19 & 37.59 & 54.25 / 68.72 / 0.6345\\
		~& CCT~\citep{CCT} & 83.22 & 80.80 & 29.47 &  70.37 & 63.16 & 49.08 & 62.68 / 75.27 / 0.7446 \\
        ~& CPS~\citep{chen2021-CPS} & 85.97  & 82.82 & 28.20 & 72.03 & 66.97 & 53.76 & 64.96 / 76.82 / 0.7693 \\
		~& LSST\citep{LSST} & 85.53 &  83.14 & 36.67 &  71.34 &  70.78 & 57.99 &  67.58 / 79.10 / 0.7757\\
		~& FixMatch~\citep{fixmatch} & 86.53 & 84.01 &36.57 & 71.26 & 69.88 & 57.38 & 67.60 / 79.34 / 0.7803 \\
		~& UniMatch~\citep{unimatch} & 84.88 & 82.75 & 34.36 &  69.87 & 66.61 & 53.03 & 65.25 / 77.51 / 0.7652 \\
        ~& DWL\citep{dwl} &  85.46 & 83.63 &  38.95 & 72.40& 70.76 & 60.33 & 68.59 / 80.24 / 0.7862 \\
         ~&Unimatchv2 ~\citep{unimatchv2} & 86.73 & 86.94 & \textbf{42.09} & \textbf{79.42 }& \textbf{81.36} & 63.33 & 73.31 / 83.49 / 0.8297 \\
            ~& Our &\textbf{86.81} &\textbf{ 87.11} & 42.05 & 78.38 & 80.93 & \textbf{65.60} & \textbf{73.48 / 83.63 / 0.8322} \\
		\midrule
		\textbf{100\%} & FullySup &88.65 & 89.94 & 51.90 & 83.40 & 84.69 & 72.83 & 78.57 / 87.32 / 0.8696\\
		\bottomrule
	\end{tabular*}
    }
     \label{tab:deepglobe}
\end{table*}
\subsubsection{Visualizatiuon Comparison}

\begin{figure*}[t]
  \centering
   \includegraphics[width=\linewidth]{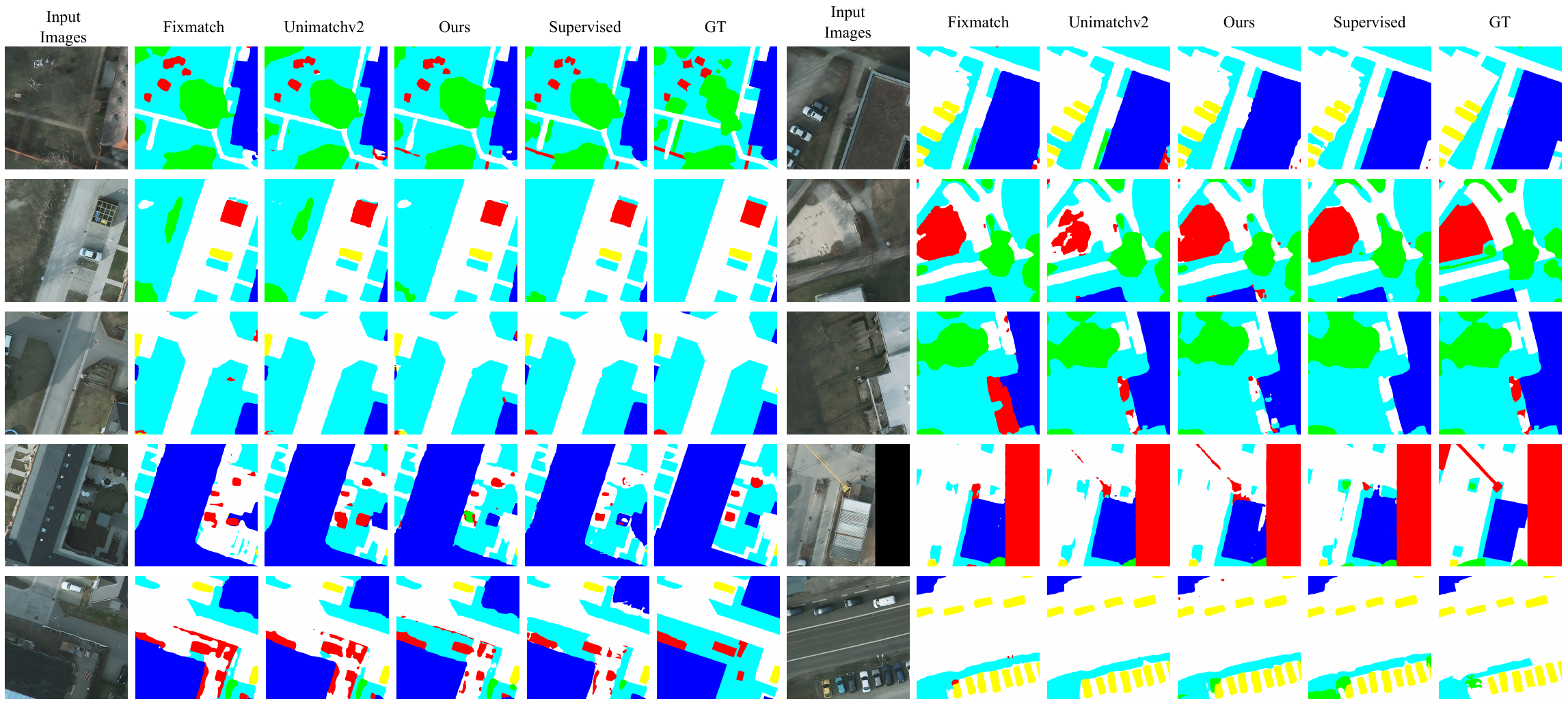}
 \caption{Qualitative comparison of different semi-supervised segmentation methods on the ISPRS Potsdam dataset. From left to right: input image, FixMatch, UniMatchv2, our method, fully supervised and grount truth.}
   \label{fig:visualization}
\end{figure*}

We provide a qualitative comparison of segmentation results generated by different methods on the ISPRS Potsdam dataset, as illustrated in Figure~\ref{fig:visualization}. Visual inspection reveals that both FixMatch and UniMatchv2, while strong baselines, exhibit noticeable errors across several categories. In particular, they frequently struggle to accurately delineate regions belonging to the \textit{Background} class (marked in \colorbox{red}{\hspace{0.8em}}), often misclassifying them or producing imprecise boundaries. 


In contrast, our RS-MTDF method demonstrates superior discriminative capability in these challenging scenarios. The segmentation results produced by RS-MTDF are visually much closer to the ground truth and the fully supervised baseline. This is particularly evident in the accurate delineation of object boundaries and the clear separation of confusing categories like \textit{Low Vegetation}~(marked in~\colorbox[rgb]{0,1,1}{\hspace{0.8em}})and \textit{Tree}~(marked in~\colorbox[rgb]{0,1,0}{\hspace{0.8em}}). This qualitative analysis further corroborates the quantitative findings, highlighting RS-MTDF's enhanced effectiveness and generalization under limited supervision.

\subsection{Ablation Studies}
\subsubsection{The Choice of Teacher Model}


To validate the effectiveness and complementarity of our selected VFMs as teacher models, we conducted experiments using various VFM configurations: DINOv2-Base, CLIP ViT-L/14, and SAM-Base. We evaluated their performance both individually and in pairwise combinations as teachers. The results, specifically focusing on the ISPRS Potsdam dataset under the 1\% label ratio, are summarized in Table~\ref{tab:chosen_teacher}.

The results unequivocally demonstrate that the combination of CLIP and DINOv2 yields the best performance, achieving an mIoU of 76.99\%. This significantly outperforms using CLIP alone (75.74\% mIoU) or DINOv2 alone (76.57\% mIoU). This observation strongly supports our hypothesis regarding the ``complementary strengths'' of these models: DINOv2 provides robust spatial awareness and local feature understanding, while CLIP contributes broad semantic understanding and global contextual priors from its vision-language pre-training. Their combined guidance leads to a more comprehensive and powerful distillation. In contrast, using SAM-Base as a sole teacher resulted in a notably lower mIoU of 74.54\%, suggesting that while SAM excels at object prompting, it may provide less effective or direct semantic supervision for pixel-wise segmentation in this semi-supervised setting compared to DINOv2 and CLIP. Even when combined with CLIP, the performance was not as strong as the DINOv2+CLIP combination.

\begin{table}
\caption{Comparison of different vision foundation models (VFMs) used as teacher models. All experiments are conducted on the ISPRS Potsdam dataset under the 1\% label ratio.}

  \label{tab:chosen_teacher}
  \centering

  \begin{tabular}{@{}l|cccc@{}}
    \toprule[1pt]
    Method & mIoU & mF1  & Kappa  \\
    \midrule
    FixMatch &75.90 & 82.42 & 0.7979 \\
    CLIP only & 75.74 & 82.21 & 0.7961\\
    DINOv2 only &  76.57 & 82.68 & 0.7955\\
     SAM only &  74.54 & 81.38 & 0.7858\\
     CLIP$+$SAM & 75.74 & 82.13 & 0.7956\\
    \midrule
    CLIP$+$DINOv2 (Ours) & \textbf{76.99} & \textbf{83.22} & \textbf{0.8004} \\

    \bottomrule[1pt]
  \end{tabular}
\end{table}

\subsubsection{Choice of Confidence Threshold ($\tau$)}

\begin{figure}[t]
  \centering
   \includegraphics[width=1.0\linewidth]{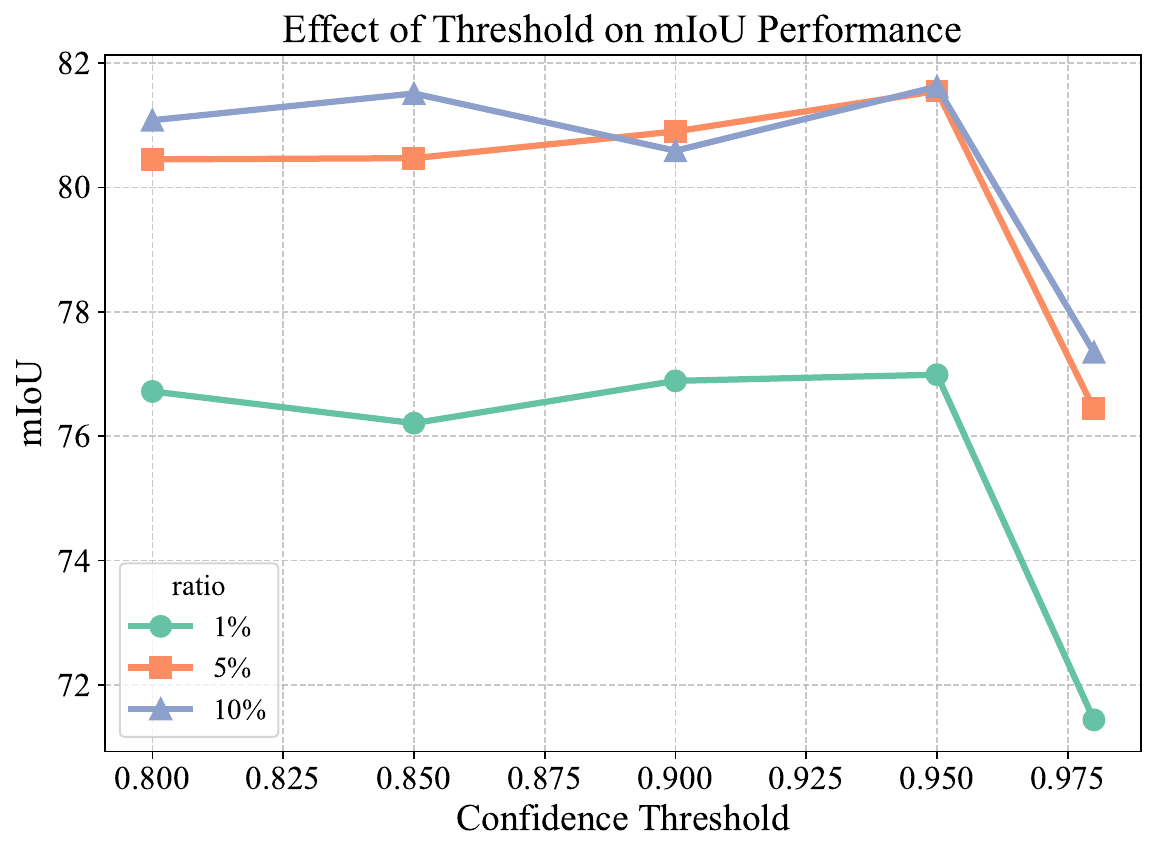}
   \caption{Ablation Study on different valus of confidence threshold in our method on the Potsdam dataset}
   \label{fig:fig2}
\end{figure}

The confidence threshold ($\tau$) plays a pivotal role in semi-supervised learning, directly influencing the quality and quantity of pseudo-labels generated by the teacher model. A higher threshold ensures higher precision but fewer pseudo-labels, while a lower threshold increases coverage but risks introducing noisy labels. To investigate the impact of this parameter, we conduct experiments with multiple threshold values (0.80, 0.85, 0.90, 0.95, and 0.98) on our method on the Potsdam dataset. The results, depicted in Figure~\ref{fig:fig2}, show the mIoU performance across these values. Our selected threshold of 0.95 consistently demonstrates the most favorable segmentation performance by striking an optimal balance between the precision (reliability) and recall (quantity) of pseudo-labels, which is crucial for stable and effective learning in low-label regimes.

\subsubsection{Contribution of Proposed Components}

To thoroughly investigate the individual and synergistic effects of our proposed teacher supervision (VFM-guided feature distillation) and multi-teacher feature fusion mechanisms, we conduct ablation studies on the DeepGlobe dataset under both 5\% and 10\% label ratios. The results are summarized in Table~\ref{tab:ablation_teacher_fusion}.

From Table~\ref{tab:ablation_teacher_fusion}, we observe the following:
\begin{itemize}
    \item \textbf{Effectiveness of VFM-Guided Feature Distillation:} Introducing teacher supervision alone (row ``Teacher: \checkmark, Fuse: \ding{55}'') leads to a noticeable improvement in mIoU, mF1, and Kappa compared to the baseline without VFM guidance. For example, at 5\% label ratio, mIoU improves from 70.89\% to 71.53\%. This clearly demonstrates the effectiveness of our multi-teacher distillation approach in enhancing the student encoder's representational capacity by transferring robust, generalizable priors from the VFMs.
    \item \textbf{Contribution of Multi-Teacher Feature Fusion:} When both teacher supervision and multi-teacher feature fusion are enabled (row ``Teacher: \checkmark, Fuse: \checkmark''), the model achieves the best performance across all metrics for both label ratios. For example, under the 10\% setting, mIoU further improves from 72.12\% (teacher supervision only) to 73.48\%, and Kappa increases from 0.8222 to 0.8322. These results confirm that the fused semantic features, enriched by VFM guidance, not only enhance the decoder input but also play a critical role in maintaining stronger spatial and semantic consistency throughout the segmentation process, which is particularly beneficial under low supervision conditions. The fusion mechanism effectively leverages the distilled VFM knowledge to refine high-level representations, leading to more accurate and coherent predictions.
\end{itemize}
These ablation findings unequivocally validate the significant contribution of each proposed VFM-guided feature distillation and multi-teacher feature fusion and highlight their synergistic effect in boosting the overall performance of RS-MTDF for remote sensing SSS




\begin{table}[ht]
\centering
\caption{Ablation study on the effect of teacher supervision and feature fusion under 5\% and 10\% label ratios on DeepGlobe Dataset.}
\begin{tabular}{cccccc}
\toprule
\textbf{Ratio} & \textbf{Teacher} & \textbf{Fuse} & \textbf{mIoU} & \textbf{mF1} & \textbf{Kappa} \\
\midrule
\multirow{3}{*}{5\%}
& \ding{55} &\ding{55} & 70.89 & 81.72 & 0.8110 \\
& \checkmark & \ding{55} & 71.53 & 82.14 & 0.8149 \\
& \checkmark & \checkmark & \textbf{71.70} & \textbf{82.24} & \textbf{0.8164} \\
\midrule
\multirow{3}{*}{10\%}
& \ding{55} & \ding{55} & 72.13 & 82.57 & 0.8239 \\
& \checkmark & \ding{55} & 72.12 & 82.62 & 0.8222 \\
& \checkmark & \checkmark & \textbf{73.48} & \textbf{83.63} & \textbf{0.8322} \\
\bottomrule
\end{tabular}
\label{tab:ablation_teacher_fusion}
\end{table}

\section{Conclusion}
\label{sec:conclusion}

In this paper, we introduced RS-MTDF, a novel and robust framework for SSS in remote sensing imagery. Our work was fundamentally motivated by the critical observation that the prevalent distribution mismatch between limited labeled data and abundant unlabeled data often severely curtails the generalization capabilities of existing semi-supervised approaches, particularly in the diverse and complex remote sensing domain. To surmount this challenge, RS-MTDF strategically leverages the powerful, generalized semantic and spatial priors embedded in VFMs. Our approach employs a multi-teacher distillation strategy, where multiple frozen VFM encoders (specifically DINOv2 and CLIP) act as expert teachers, providing feature-level guidance to align the student encoder's representations with their robust, high-quality features. Furthermore, to fully capitalize on the rich knowledge acquired through distillation, we proposed a multi-teacher feature fusion component, seamlessly integrating these translated, VFM-enhanced features directly into the student's decoder pathway, leading to significantly more discriminative and accurate segmentation predictions. Extensive experiments conducted on three challenging benchmark remote sensing datasets ISPRS Potsdam, LoveDA, and DeepGlobe Land Cover—unequivocally demonstrate the effectiveness and robustness of RS-MTDF. Our method consistently achieves state-of-the-art performance, notably excelling under extremely low-label regimes, where existing approaches often falter. Overall, RS-MTDF provides a simple yet effective paradigm for integrating the unparalleled generalization power of large-scale pre-trained VFMs into traditional semi-supervised learning pipelines, offering a practical solution to the persistent data scarcity and distribution mismatch problems in remote sensing semantic segmentation and opening up exciting new directions for future research.

\section{Limitation and Future Work}
\label{sec:limitation}


Our proposed RS-MTDF framework, while demonstrating state-of-the-art performance in remote sensing SSS, does present certain limitations, primarily pertaining to its training efficiency and resource requirements.

A significant strength of RS-MTDF is that the introduction of multiple VFMs and additional translator/fusion modules is exclusively confined to the training phase. Critically, these components are not required during inference or deployment. Consequently, our framework incurs zero extra computational cost at inference time, thereby maintaining the same efficiency as traditional encoder-decoder models during real-world application. This makes RS-MTDF highly practical for deployment in scenarios where computational resources are constrained post-training.

However, the inclusion of multiple VFM-based teacher models and their associated feature alignment modules (translators and linear projectors) inherently increases the memory usage and computational load during the training process. This elevated demand for GPU memory and processing power may pose considerable challenges for researchers attempting to train RS-MTDF on resource-limited hardware or when scaling the method to even larger remote sensing datasets than those evaluated in this study. The concurrent loading of multiple VFM models, student models, and their respective feature maps during distillation contributes to this heightened resource consumption.

For future work, addressing these training-time limitations will be a key focus. We envision several promising avenues:
\begin{itemize}
\item More Lightweight Teacher Configurations: Exploring the use of smaller, distilled versions of VFMs (e.g., Tiny or Small variants) as teachers, or even task-specific fine-tuned VFMs that are more compact, could significantly reduce memory footprint.
\item Sequential or Asynchronous Distillation: Instead of concurrent loading, exploring sequential distillation from multiple teachers or asynchronous update mechanisms might alleviate peak memory usage.
\item Optimized Training Architectures: Researching novel architectural designs for feature translators and fusion modules that are inherently more resource-efficient during training could further enhance scalability.
\end{itemize}
By exploring these directions, future iterations of RS-MTDF or similar VFM-guided SSS frameworks can achieve even greater training efficiency, broadening their applicability to a wider range of computational environments and larger-scale remote sensing projects.

\section*{Declaration of Competing Interest}

Declaration of Competing Interest The authors declare that they have no known competing financial interests or personal relationships that could have appeared to influence the work reported in this paper.

\section*{Acknowledgements}

This work is partially supported by the National Key R\&D Program of China (2021ZD0112902), and China NSFC projects under contract 62272375, 12226004.

\printcredits

\bibliographystyle{cas-model2-names}

\bibliography{ref}





\end{document}